\documentclass[sigconf, natbib=true]{acmart}
\AtBeginDocument{%
  }

\copyrightyear{2025}
\acmYear{2025}
\setcopyright{acmlicensed}\acmConference[SIGIR '25]{Proceedings of the 48th International ACM SIGIR Conference on Research and Development in Information Retrieval}{July 13--18, 2025}{Padua, Italy}
\acmBooktitle{Proceedings of the 48th International ACM SIGIR Conference on Research and Development in Information Retrieval (SIGIR '25), July 13--18, 2025, Padua, Italy}
\acmDOI{10.1145/3726302.3730068}
\acmISBN{979-8-4007-1592-1/2025/07}

\usepackage[T1]{fontenc}
\usepackage[utf8]{inputenc}

\usepackage{mathtools} 
\usepackage{amsmath} 
\usepackage{graphicx,color}
\usepackage{tabularx}
\usepackage{booktabs}
\usepackage{multirow}
\usepackage{ragged2e} % For \RaggedRight
\usepackage{mdframed}
\usepackage{algorithm}
\usepackage{algpseudocode}
\usepackage[subtle]{savetrees}

\usepackage{listings}

\definecolor{lightgray}{gray}{0.85}

\author[H. Azarbonyad]{Hosein Azarbonyad}
\authornote{Equal Contribution.}
\affiliation{%
  \institution{Elsevier}
  \city{Amsterdam}
  \country{The Netherlands}
}
\email{h.azarbonyad@elsevier.com}

\author[Z. Zhu]{Zi Long Zhu}
\authornotemark[1]
\affiliation{%
  \institution{Elsevier}
  \city{Amsterdam}
  \country{The Netherlands}
}
\email{z.zhu@elsevier.com}

\author[G. Cheirmpos]{Georgios Cheirmpos}
\authornotemark[1]
\affiliation{%
  \institution{Elsevier}
  \city{Amsterdam}
  \country{The Netherlands}
}
\email{g.cheirmpos@elsevier.com}

\author[Z. Afzal]{Zubair Afzal}
\affiliation{%
  \institution{Elsevier}
  \city{Amsterdam}
  \country{The Netherlands}
}
\email{zubair.afzal@elsevier.com}

\author[V. Yadav]{Vikrant Yadav}
\affiliation{%
  \institution{Elsevier}
  \city{Amsterdam}
  \country{The Netherlands}
}
\email{vikrant4.k@gmail.com}

\author[G. Tsatsaronis]{Georgios Tsatsaronis}
\affiliation{%
  \institution{Elsevier}
  \city{Amsterdam}
  \country{The Netherlands}
}
\email{g.tsatsaronis@elsevier.com}

\renewcommand{\shortauthors}{Hosein Azarbonyad et al.}

\begin{document}

\title{Question-Answer Extraction from Scientific Articles  Using Knowledge Graphs and Large Language Models}


\begin{abstract}
When deciding to read an article or incorporate it into their research, scholars often seek to quickly identify and understand its main ideas. In this paper, we aim to extract these key concepts and contributions from scientific articles in the form of Question and Answer (QA) pairs. We propose two distinct approaches for generating QAs.
The first approach involves selecting salient paragraphs, using a Large Language Model (LLM) to generate questions, ranking these questions by the likelihood of obtaining meaningful answers, and subsequently generating answers. This method relies exclusively on the content of the articles.
However, assessing an article's novelty typically requires comparison with the existing literature. Therefore, our second approach leverages a Knowledge Graph (KG) for QA generation. We construct a KG by fine-tuning an Entity Relationship (ER) extraction model on scientific articles and using it to build the graph. We then employ a salient triplet extraction method to select the most pertinent ERs per article, utilizing metrics such as the centrality of entities based on a triplet TF-IDF-like measure. This measure assesses the saliency of a triplet based on its importance within the article compared to its prevalence in the literature.
For evaluation, we generate QAs using both approaches and have them assessed by Subject Matter Experts (SMEs) through a set of predefined metrics to evaluate the quality of both questions and answers. Our evaluations demonstrate that the KG-based approach effectively captures the main ideas discussed in the articles. Furthermore, our findings indicate that fine-tuning the ER extraction model on our scientific corpus is crucial for extracting high-quality triplets from such documents.
\end{abstract}




\maketitle
\def\thefootnote{*}\footnotetext{These authors contributed equally to this work}
\def\thefootnote{\arabic{footnote}}
\section{Introduction}
Scientific document users (researchers, students, or scholars) usually deal with a large number of articles in their daily work, making it increasingly challenging to efficiently identify and comprehend the core contributions and ideas within this vast literature. This challenge is particularly pronounced in the context of academic research, where a thorough understanding of the relevant literature is critical to inform and advance new research endeavors. Traditional methods of literature review and analysis, which often involve reading and annotating large volumes of documents, can be time-consuming and labor-intensive, posing a significant barrier to efficient knowledge acquisition and synthesis \cite{jeschke2019knowledge, carver2013identifying, plaven2017readability}.
To address these challenges, automatic reading assistant tools have been developed, offering innovative solutions to streamline the literature review process. Techniques such as text summarization aim to condense large texts into shorter and more manageable summaries that retain the core information \cite{cohan2018scientific, mishra2022scientific, agarwal2011scisumm}. Most (if not all) approaches aim to generate unstructured summaries (abstractive or extractive). However, in the context of long documents, grasping the main ideas based on a free-text summary could be challenging. 

In this paper, our aim is to streamline the process of understanding academic articles by generating Question-Answer (QA) pairs that encapsulate the main ideas and contributions of scientific articles. This approach aims to provide researchers with concise and accessible summaries in the form of QA pairs, enabling them to quickly ascertain the relevance and significance of an article to their research without needing to delve deeply into the full text initially.
QA pairs provide targeted answers to specific questions, can clarify complex concepts by breaking down information into manageable, straightforward questions and answers, and encourage active engagement with the material. 
Moreover, such QAs can be directly used to train and evaluate the performance of question answering models in the scientific context.

Our research introduces two approaches for generating QAs from full-text articles: a \textbf{C}ore \textbf{C}ontext-based \textbf{Q}A \textbf{G}eneration approach (CCQG) and a \textbf{K}nowledge \textbf{G}raph (KG)-based approach. CCQG leverages key elements of an article—such as the title, abstract, and keywords—to identify and extract paragraphs with high similarity to these elements. These selected salient paragraphs serve as the foundation for generating questions using a Large Language Model (LLM). The generated questions are then aggregated and ranked based on the likelihood of having high-quality answers. The same LLM is subsequently used to generate answers for these questions, with the quality of the QAs being evaluated based on criteria such as length, grammatical correctness, and overall usefulness.

Recognizing the importance of contextual relevance and novelty in the scientific literature, our second approach utilizes a KG constructed from a set of articles to assess novelties of articles against the literature. This KG-based approach involves extracting Entity Relationships (ERs) from key documents using an LLM, which are then used to fine-tune a triplet extraction model. The triplet extraction model is used to construct the KG. From this KG, we extract the most salient ERs to each article, using a TF-IDF-like measure adapted for triplet frequency. These salient triplets are then fed into an LLM to generate questions, followed by a similar ranking and answer generation process as in the CCQG approach.

To evaluate the effectiveness of our proposed QA generation methods, we conduct an evaluation involving Subject Matter Experts (SMEs) who assess the quality of the generated QAs across a set of metrics, including relevance, usefulness, specificity, and factuality \cite{es2023ragas, yu2024evaluation}. Our results demonstrate that the KG-based approach, in particular, produces QAs that more comprehensively cover the main content of the articles, as indicated by higher SME scores compared to the CCQG method.
By providing an efficient means to distill and present the essential content of academic articles, this approach holds a significant potential to enhance the accessibility and utility of scientific literature, thereby supporting researchers in their efforts to quickly and effectively identify pertinent research insights. 
%The generated QAs for 20,000 articles in the Computer Science domain are currently publicly available on a popular academic articles database.

Our main contributions in this paper are as follows:
\begin{enumerate}
    \item We propose the task of generating high-quality Question-Answer (QA) pairs for scientific articles to summarize their main ideas and contributions. These QA pairs offer an efficient and engaging way for researchers to quickly grasp the relevance and significance of an article.
    \item We introduce two approaches for QA generation including a Knowledge Graph (KG)-based method that extracts Entity Relationships to represent the main ideas of articles in the context of the literature.
    \item We propose a robust evaluation framework that uses metrics such as relevance, usefulness, and accuracy. QA quality is assessed by Subject Matter Experts (SMEs) and an "LLM-as-a-judge" paradigm, demonstrating the effectiveness of our KG-based approach in generating high-quality QAs.
\end{enumerate}
%The remainder of this paper is organized as follows:
%In Section \ref{relatedWork}, we review the related literature around QA generation and scientific document summarization.
%Section \ref{methods} describes the details of the two methods designed for QA generation. 
%In Section \ref{experimentalSetup}, the details of the experiments are explained.
%We preset the results of different QA generations methods in Section \ref{results}.
%Section \ref{limitations} discusses the main limitations of the current study and the designed methods.
%Finally, in Section \ref{conclusion}, we conclude the paper and discuss a set of possible future directions.

\section{Related Work}
\label{relatedWork}

Our QA generation work touches upon several related work from different aspects: QA generation, document summarization, and knowledge graph construction.
%In this section, we review the related work on these areas.

\subsection{QA generation}
%Advancements in question generation and answering (QA) systems have focused on enhancing the diversity and quality of synthetic training data to improve QA performance. 
The simplest approach to generating QAs is the use of rule-based methods which has been shown to be effective in extracting QAs in the educational domain \cite{rao2022generating}. \citet{sultan2020importance} have tried to use the $top-p$ nucleus sampling to promote textual diversity in QA generation. Several studies have tried to integrate models of question generation and answer extraction \cite{alberti2019synthetic, tang2017question}. \citet{duan2017question} trained neural question generation methods from large-scale QA pairs integrating retrieval-based and generation-based approaches demonstrating significant QA improvements. Similarly, a Retrieve-Generate-Filter technique has been introduced to create counterfactual evaluation and training data \cite{paranjape2021retrieval}. 
Most of these QA extraction methods focus on extracting factoid questions from a short context (paragraphs). In this paper, we aim to extract QAs that can help researchers understand full-text articles in the scientific context.

\subsection{Document Summarization}
Our generated QAs aim to extract QAs that help researchers understand the main topics and novelties of articles. Therefore, our approach has some commonalities with text summarization. 
Most of the research on summarization has focused on addressing key challenges such as informativeness and factual consistency by incorporating external knowledge or advanced models. 
\citet{ji2021skgsum} tried to integrate semantic KGs into the summarization model to capture the relationships between sentences and entities. To ensure factual consistency, entity-level consistency metrics and summary-worthy entity classification models are integrated into the training process \cite{nan2021entity}. 
%The SENECA framework employs an entity-aware content selection module and an abstract generation module to enhance the coherence, conciseness, and clarity of summaries \cite{sharma2019entity}. 
The combination of graph-based syntactic representations and centrality metrics has also been shown to be effective in extractive summarization \cite{litvak2008graph}. 
In the context of scientific document summarization, the use of citations and scientific discourse structures to contextualize and summarize articles has been shown to be effective in generating high-quality summaries \cite{cohan2018scientific, mishra2022scientific, agarwal2011scisumm}. The Information Bottleneck principle has been extended to document summarization, achieving a high effectiveness in covering more content aspects in scientific documents \cite{ju2021leveraging}. \citet{cachola2020tldr} introduced the TLDR task, the SciTLDR dataset, and a learning strategy called CATTS to generate concise summaries. 
All of these methods aim to extract or generate a summary for articles. Our goal is to extract research questions and their answers that can reflect not only the main ideas of the articles but also their novelties with respect to the literature.

\subsection{Knowledge Graph Construction}
Transforming unstructured text into a KG has been a focus of several studies. \citet{martinez2018openie} integrated Open Information Extraction with grammatical structures to produce coherent, proposition-based KG facts.
%, demonstrating improved understanding and representation of extracted information. 
To enhance scalability and reduce redundancy, \citet{clancy2019knowledge} tried to map the extracted triples to a homogeneous namespace such as DBpedia. 
\citet{stewart2020seq2kg} proposed a scalable open source platform to extract mentions and relations, supporting a variety of applications including fact verification and training data extraction. The Seq2KG model constructs KGs from scratch by jointly generating triples and resolving entity types \cite{mehta2019scalable}. 

Relation Extraction (RE) is the core technique used for creating KGs \cite{martinez2018openie,clancy2019knowledge,stewart2020seq2kg,huguet-cabot-navigli-2021-rebel-relation,wadhwa2023revisiting,giorgi2022sequence,dagdelen2024structured,ren2022sim,xu2022emr}. 
The REBEL model performs end-to-end relation extraction for over 200 different types of relations, achieving state-of-the-art results in various benchmarks \cite{huguet-cabot-navigli-2021-rebel-relation}. Similarly, \citet{wadhwa2023revisiting} used LLMs and demonstrated the effectiveness of treating RE as a seq2seq task, achieving high performance even with few-shot learning.
Extending these techniques to document-level relation extraction, the DocRE model integrates entity extraction, coreference resolution, and relation extraction in a single framework \cite{giorgi2022sequence}. In specialized domains, fine-tuning LLMs for joint named entity recognition and relation extraction has proven to be effective, enabling the extraction of structured knowledge from scientific texts in various formats \cite{dagdelen2024structured}. 
Inspired by these advancements, we use the REBEL model and fine-tune it on the scientific data to extract entities and relationships and build a KG of scientific articles.

\section{Question-Answer Pair Generation}
\label{methods}
In this section, we describe two different methods for generating QA pairs based on full-text articles. 
Each method takes an article $D$ as input, where each article contains full text, title, abstract, and a set of keywords. 
For each article $D$, each QA generation method generates a tuple of QAs: $QA_D={\{(q_i, a_i, p_i)\}}_{i=1}^{l}$, where $q_i$ is a question, $a_i$ is an answer, and $p_i$ is a set of paragraphs used to generate the answer with respect to the question. 

\subsection{Core Context-based QA Generation} 
\label{contextBased}
In this section, we describe the Core Context-based QA Generation approach (CCQG) and the steps taken to generate a QA set per article. 
The main idea behind this approach is to select the salient paragraphs in the article and generate QAs based on them. 
To select paragraphs and generated QAs, this method uses the following steps:
salient paragraph selection, question prompting, filtering and ranking, and answer generation and filtering. 

\subsubsection{Salient Paragraph Selection}
In this step, the objective is to extract the most salient paragraphs from the article that discuss key points of the article. 
We first use the XML structure of the articles and extract the natural paragraphs as written by the authors. 
Then, we use a target proxy to rank the paragraphs according to the information they contain. 
The proxy structure is the concatenation of title, abstract, and keywords of the article.
The paragraphs are ranked on the basis of the similarity of their representations to the representation of the proxy (we use a sentence transformer to extract the representations).
We then select the paragraphs with a similarity higher than a threshold to the proxy as the salient paragraphs. 

\subsubsection{Question Generation} 

Once the salient paragraphs are selected, we use an LLM to generate a set of questions based on each paragraph
%We prompt the LLM with a set of instructions on how to generate questions 
(the prompt can be seen in Figure \ref{fig:prompt-q}). 
The prompt structure is composed of the relevant paragraphs, along with the title, text, and keywords for a focused scope. 
For each relevant paragraph, we aim to generate a set of questions that are about the core ideas or applications that may be discussed in the article. 
We aggregate all questions generated across all paragraphs as candidate questions for the final QA per article and try to rank them.
%Before rankig, we de-duplicate questions by grouping high similarity questions together and selecting a question per group. 

\begin{figure}[hbt!]
\begin{lstlisting}
You are a scientific expert AI assistant, and your task is to generate questions for articles with the aim to help researchers, such as scientists, scholars, students, understand the content of the articles. The user provides you with a set of paragraphs, title, and keywords of a scientific article given in triple backticks. You need to generate questions based on the provided paragraphs.

### INSTRUCTIONS ###
1.Take your time to read the title and keywords carefully to understand the main topic of the article.
2.Craft questions that focus on exploring the article's scientific value, emphasizing main ideas and contributions, while maintaining simplicity in language structure and conciseness. Avoid queries that delve into specific sections, paragraphs, references, tables, figures, methods, or authors.
3.Ensure that each question pertains to the content in a paragraph. The number of questions generated for each paragraph may vary.
4.Ensure your questions are moderately specific without being too detailed or overly generalized. 
5.Employ a variety of interrogative adverbs and auxiliary verbs, including but not limited to 'How', 'When', 'Why', 'Does'.Refrain from using the same one repeatedly to ensure diversity in questioning. 
6.If any abbreviation is used, make sure they also provide the longer form (extension) of the abbreviation.

### TO AVOID ### 
1.Ensure your questions are moderately specific without being too detailed or overly generalized. 
2.Do not generate questions that include references to specific sections, paragraphs, citations, tables, figures, methods, equation, authors.
3.Do not use the words "paper", "article", "methods", "proposed", or similar words in questions.
Take a step back and review the result.

### CONTEXT ### 
"""
    Title of article: {TITLE}
    Keywords of article: {KEYWORDS}
    
    Paragraph 1:  {para1_text}
    Paragraph 2:  {para2_text}
    ...

Generate questions for each of the paragraphs. Provide the results in a JSON struct and nothing else.
"""
\end{lstlisting}

\caption{Question generation prompt for the CCQG method.}
\label{fig:prompt-q}
\end{figure}

\noindent
\textbf{Question Ranking and Filtering}
Question ranking focuses on prioritization of questions given for generation and consists of two phases: \textbf{greedy ranking} and \textbf{supplementary ranking}. The details of the ranking method can be found in Algorithm \ref{alg:cap}.
Greedy ranking starts by aggregating the similarity scores of all the questions $\mathcal{Q}$ with all salient paragraphs $\mathcal{P}_{salient}$.
The questions are iterated by highest aggregated score order at every step.
A question is paired with the first available paragraph from the set $\mathcal{P}_{salient}$ via a similarity score function $sc$. If the score is higher than a threshold $th$ then $p$ is assigned to the question $q \in \mathcal{Q}$. Subsequently, we remove paragraph $p$ from $\mathcal{P}$.
The process stops when all questions are paired with up to $K=3$ paragraphs or there are no paragraphs left to be paired with the questions. 
Questions that are not assigned to any paragraph are ranked through a supplementary ranking process against full text of the article. 
In this step, questions are sorted based on their aggregated similarity score to all paragraphs $\mathcal{P}_{rest}$ of the article (excluding salient paragraphs), and given an ascending rank starting from the position after the last question from the greedy ranking. 
All questions from the supplementary ranking are assigned to the paragraph $p_o$ they originated from along with the $K=3$ paragraphs that are extracted from the most contextually similar paragraphs to the question from the whole article.
\begin{algorithm}
\caption{Two step question ranking}\label{alg:cap}
\begin{algorithmic}
\Require $\mathcal{Q}, \mathcal{P}_{salient}, \mathcal{P}_{rest}, th, sc, K$

\State $\mathcal{S} \gets \{\}$ \Comment ordered set with paired question and paragraphs
\While{$\mathcal{Q} \neq \emptyset$ and $\mathcal{P}_{salient} \neq \emptyset$} \Comment greedy ranking
\State$q, p_1, ... , p_{ K}= sc(\mathcal{Q}, \mathcal{P}_{salient}, th, K)$
\State$\mathcal{S} \gets \mathcal{S} \cup (q, p_1, ... , p_{ K})$
\State$\mathcal{Q} \gets \mathcal{Q} \setminus \{q\}$
\State$\mathcal{P}_{salient} \gets \mathcal{P}_{salient} \setminus\{p_1, ... , p_{ K}\}$
\EndWhile
\While{$\mathcal{Q} \neq \emptyset$} \Comment supplementary ranking
\State$q, p_o, p_1,... , p_{ K} = sc(\mathcal{Q}, \mathcal{P}_{rest}, th, K)$
\State$\mathcal{S} \gets \mathcal{S} \cup (q, p_o,p_1, ... , p_{ K})$
\State$\mathcal{Q} \gets \mathcal{Q} \setminus \{q\}$
\EndWhile
\State \Return $\mathcal{S}$
\end{algorithmic}
\end{algorithm}
%
%\noindent
%\textbf{Question Filtering}
After ranking, we filter out questions with specific patterns in them such as mentions of authors, tables, or figures \footnote{The motivation for this filtering step is that the eventual generated QAs should be understandable without prior knowledge of the article's contents.}.
To increase the coverage of QAs to the whole article, we limit the number of questions originated from each section to two questions. 

\subsubsection{Answer Generation}
After question ranking and assigning paragraphs to the questions, we use an LLM to generate an answer for each question based on its paragraphs.
The answer prompt is a two-part prompt, where both the answer and the references to the answer have to be generated at once (the prompt can be found in Figure \ref{fig:prompt-a}). 
As a context, we provide the set of relevant paragraphs for the question, making it a question-oriented prompt. 
References are supporting sentences from the relevant paragraphs which are used to generate the answer. 
%This makes it possible to provide pointers to the full article where the answer is originated from.

\begin{figure}
\begin{lstlisting}
You are a Scientific Question Answering System with Source Attribution and your task it to answer questions based on provided paragraphs. You are provided a set of information regarding a scientific article. This information includes the title, keywords and a set of paragraphs belonging to the article. The question is directly associated with the information given.Make sure to study them thoroughly.

### INSTRUCTIONS ###
1. Do not try to make up an answer. The answer should only be based on the provided paragraphs. If there is no relevant information in the paragraphs or if the question cannot be answered based on the paragraphs, then the answer should be "Irrelevant context to the question. No answer provided."
2. Ensure the answer is self-contained, requiring no reference to other parts of the article.
3. Provide concise responses, between 40 to 100 words.
4. Avoid specific references to sections, paragraphs, citations, tables, figures, equations, or authors.
5. If abbreviations are used, ensure to include the corresponding full form (extension) of the abbreviation.

### OTHER GUIDELINES ###
1.	Copy the sentences and number from the original paragraphs provided by the user, to help them identify where the answer came from.
2.	Make sure you do not change the text of the paragraphs and provide it as is.
3.	Ensure that sentences are grammatically correct and complete. 
4.	Maintain consistent tense throughout the answer sentences.
5.	Use clear and easily understandable language. 

Take a moment to review and ensure the response adheres to the guidelines from above.

### CONTEXT ### 
"""
    Title of article: {TITLE}
    Keywords of article: {KEYWORDS}
    
    Paragraph 1:  {para1_text}
    Paragraph 2:  {para2_text}
    ...
"""
Question : {question}
Provide the results in a JSON struct and NOTHING else. JSON format -> {'answer': answer_text, 'sentences' : [(sentence_text, paragraphs), ...]}

\end{lstlisting}
\caption{Answer generation prompt for the CCQG and KG-based methods.}
\vspace{-10pt}
\label{fig:prompt-a}
\end{figure}

\noindent
\textbf{Answer Filtering}
We use SelfCheckGPT \cite{manakul2023selfcheckgpt} with the NLI approach to filter out answers with any possible hallucinations in them. 
This involves checking the generated answers with the source paragraphs and removing any answer which is detected to contain any hallucination.
We use similar post-processing as for the questions to remove any QA where the answer contains mentions of specific entities such as authors or sections.

\subsection{Knowledge Graph-based Approach}
\begin{figure*}[t]
\centering
\includegraphics[width=0.8\textwidth]{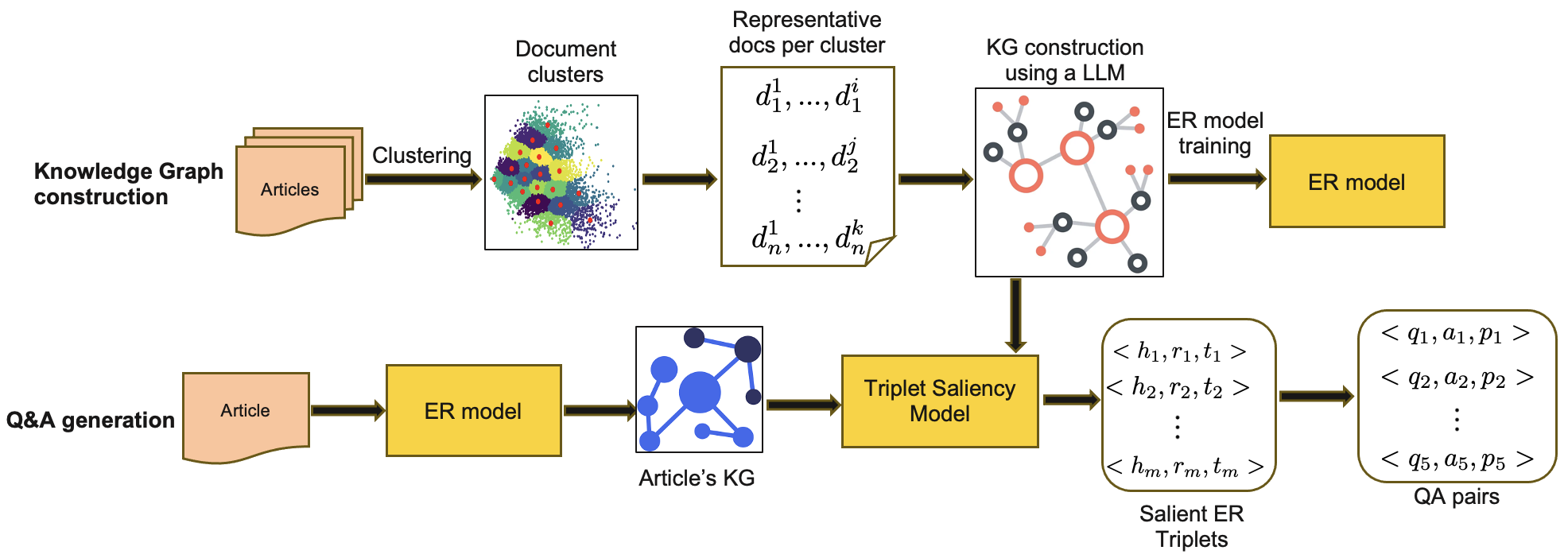}
\vspace{-0.5\baselineskip}
\caption{The KG construction and KG-based QA generation pipeline. An ER extraction model is trained by using a set of ERs extracted from a corpus of seed articles. Then, the model is used to build a small KG per article. A triplet saliency score is computed per triplet in a given article (by comparing article's KG with the KG constructed for the whole corpus) and salient triples are used to generate QAs.}
\label{fig:kg}
\end{figure*}

In the KG-based approach, we aim to generate QAs based on salient entities and their relationships in articles. 
For this purpose, we build a KG to find salient ER triplets in a given article. 
By leveraging the salient triplets and their origin in the given article, we generate questions that focus on more novel aspects of the article. Figure \ref{fig:kg} shows an overview of this approach for generating QAs.
In the following sections, we explain each step in detail.

\subsubsection{Triplet Extraction Model}
To construct the KG, we first need to extract ER triplets from each article. 
An ER triplet consists of two entities, the head (subject) $h$ and the tail (object) $t$, and their relationship $r$, which we denote as: $(h, r, t)$. 
For triplet extraction, we employ REBEL, which is a BART-based model that extracts ER triplets end-to-end with a chunk of text as input. 
We fine-tune the REBEL on our corpus of scientific articles. 

To create the dataset for fine-tuning, we select a small fraction\footnote{We opted for 5\%, as this gave us enough examples for training the model, while also keeping the cost of annotating the data low.} of the corpus. To ensure that the dataset is representative of the overall corpus and diverse, we select individual documents using a clustering-based algorithm. 
% To create the dataset for fine-tuning, we select a small fraction of the corpus (5\% documents from the initial corpus) as representatives. 
We first use TopicBert \cite{grootendorst2022bertopic} to create different clusters on different topics in our corpus. 
Within each topic cluster ${c \in C}$ (where $C$ is the number of extracted topics), we select $d_{c} = 0.05 \times \lvert {c} \rvert$  documents as representatives. 
The representatives are chosen by performing the KMeans clustering method with $d_{c}$ as centroids and choosing the articles that are closest to the centroids as the representatives $\mathcal{D}_c$. The representative corpus is then created by $\mathcal{D} = \bigcup_{c\in C} \mathcal{D}_c$.
To create a text-to-ER triplet dataset with $\mathcal{D}$, we feed the articles in chunks to an LLM to generate the ER triplets for each article.
We then use the (chunk, triplet) pair to fine-tune the REBEL model as our triplet extraction model.

\subsubsection{Knowledge Graph construction}
After extracting the ER triplets using REBEL, we construct the KG, where we keep the following information: (1) frequency of each entity per article and (2) frequency of each ER triplet per article. We construct and host the KG on a graph database\footnote{We use Neo4j as our graph database.}.

\subsubsection{Triplet Saliency}
As mentioned above, the goal is to select salient triplets and to use them to generate questions that focus on novel or interesting parts of a given article. For this purpose, we propose the salient triplet extraction method, which uses the following saliency measure for a triplet $T$, which consists of two components. The first component is defined by the following functions,
\begin{align}
    S_h &= \text{\small tf-idf}_{\text{entity}}(h) \cdot \text{\small pagerank}(h)\\
    S_t &= \text{\small tf-idf}_{\text{entity}}(t) \cdot \text{\small pagerank}(t)\\
    S^{graph}_T &= \min(S_h, S_t) \cdot  \text{\small tf-idf}_{\text{triplet}}(T)
\end{align}
where the term frequency (tf) in $\text{\small tf-idf}_{\text{entity}}(e)$ is calculated by taking the frequency of entity $e$ in the triplets that appear in the given article. The inverse document frequency (idf) is calculated based on the number of articles containing the entity $e$ (based on the triplets). For $\text{\small tf-idf}_{\text{triplet}}(T)$, tf is the frequency of the triplet $T$ in the given article, and idf is calculated based on the number of articles that contain $T$.

In the second component, we calculate the semantic similarity of a triplet $T$ with article metadata (title, keywords and abstract):
\begin{align}
    S^{semantic}_T &= \text{sem\_sim}(T, M_D)
\end{align}
where $M_D$ is the concatenation of the metadata of article $D$. We use a sentence-transformer to calculate the similarity. 
The combined saliency measure is then given by:
\begin{align}
    S_T = S^{graph}_T \cdot S^{semantic}_T
\end{align}
The motivation behind $S_{T}^{graph}$ is that we can retrieve triplets from the article with a tf-idf-like importance measure, but with additional structural information provided by the graph. This way, we rank triplets higher that are more novel conditioned on the entire corpus. $S_{T}^{semantic}$ is to keep the triplet related to the main topics of the article.

\subsubsection{Question and answer generation}
After we have ranked all the triplets from a given article, the next step is to prompt the LLM for question generation. Here, we first group consecutive triplets if they have the same head $h$ and tail $t$, creating groups of ranked triplets.
Then we take the top $m=10$ groups of triplets which are used separately in the prompt to generate the questions. 
The prompt structure is the same as the one used for the CCQG method, with the addition of triplets to the prompt (Figure \ref{fig:prompt-q-kg}).
%View the prompt in section \ref{sec:promptsappendix}. 
Next, we rank the questions using the same question ranking method and generate answers using the same answer generation method as the CCQG method. Except, we allow a paragraph to be salient for multiple questions during the question ranking phase. This difference in implementation is due to the questions not having a clear origin paragraph as in the CCQG method.

\begin{figure}[hbt!]
\begin{lstlisting}
 You are a scientific expert AI assistant, and your task is to generate questions for articles with the aim to help researchers, such as: scientists, scholars, students, understand the content of the articles. The user provides you with the following context as INPUTS: Entity Relationship (ER) triplets, title, keywords of a scientific article given in triple backticks. You need to generate questions based on the provided triplet and its associated paragraphs.

An ER triplet consists of: head, relationship, and tail, signifying the relationship between two entities, the head and tail.
            
### INSTRUCTIONS ###    
1.	Take your time to read the title and keywords carefully to understand the main topic of the article. 
2.  Base the questions on the ER triplets.
3.	Craft questions that focus on exploring the article's scientific value, emphasizing main ideas and contributions, while maintaining simplicity in language structure and conciseness. Avoid queries that delve into specific sections, paragraphs, references, tables, figures, methods, or authors. 
4.	Generate one or atmost two questions per ER triplet. 
5.	Ensure your questions are moderately specific without being too detailed or overly generalized.  
6.	Employ a variety of interrogative adverbs and auxiliary verbs, including but not limited to 'How', 'When', 'Why', 'Does', ... Refrain from using the same one repeatedly to ensure diversity in questioning.  
7.	If any abbreviation is used, make sure they also provide the longer form (extension) of the abbreviation. 
8.  Provide the results in a JSON struct 

### TO AVOID ### 
1.	Ensure your questions are moderately specific without being too detailed or overly generalized. 
2.	Do not generate questions that include references to specific sections, paragraphs, citations, tables, figures, methods, equation, authors. 
3.	Do not use the words "paper", "article", "methods", "proposed", in questions. 
Take a step back and review the result and make sure it is proper. \n\n  

### INPUTS ###
    ER Triplets: {TRIPLETS}
    Title of article: {TITLE}
    Keywords of article: {KEYWORDS}
    Paragraph 1:  {para1_text}
    Paragraph 2:  {para2_text}
    ...
### REMINDERS ###
   - Generate questions based only on the ER triplets.
   - Provide the results in a JSON struct and only the questions.
   """
\end{lstlisting}
\caption{Question generation prompt for the KG-based method.}
\label{fig:prompt-q-kg}
\end{figure}

\section{Experimental Setup}
\label{experimentalSetup}

We evaluate the performance of our proposed QA generation pipelines by means of automatic and human evaluation. Our main research questions are: (1) How effective are the proposed approaches in generating high-quality QAs which capture the main topics of articles? (2) Can we reliably fine-tune a triplet extraction model to extract triplets from a set of scientific articles?
(3) How does the proposed salient triplet extraction method perform in capturing salient triplets in articles? 
\textbf{RQ1} is concerned with the quality of QAs generated using different pipelines. To answer RQ1, we generate QAs for a set of articles and have them evaluated by SMEs on several metrics. We further measure the quality of the QAs using an LLM on the same set of metrics. The results of these experiments are presented in Section \ref{resultsRQ1}.
To answer \textbf{RQ2}, we evaluate the quality of extracted triplets by the fine-tuned REBEL model on a set of 50 articles and compare the extracted triplets with those extracted by the vanilla REBEL model. 
The results of this experiment are described in Section \ref{resultsRQ2}
To answer \textbf{RQ3}, we use our salient triplet extraction method to extract salient triplets from a set of articles and compare the performance of this method with a set of baselines in the salient triplet extraction task. The results of experiments related to RQ3 are reported in Section \ref{resultsRQ3}.

\subsection{Datasets}
\label{datasets}
We focus on the two scientific domains: Computer Science and Life Sciences. Per domain, we select 20,000 articles (published in Elsevier's ScienceDirect journals\footnote{\href{https://www.sciencedirect.com/}{https://www.sciencedirect.com/}}) from the domain to build and evaluate our QA generation methods \footnote{We use \href{https://pypi.org/project/lxm/}{lxml} package to parse the articles and extract their paragraphs.}.
To evaluate the quality of QAs, we select 20 articles per domain and extract top-5 QAs from them. Then, we have these QAs evaluated by SMEs. We also evaluate the quality of these QAs using LLMs as judges. 
Each QA is evaluated by two SMEs (for all metrics) on a scale of 1-3. The value of each metric is calculated by summing both SME scores (for the metric) over all QAs and dividing it by the maximum possible score. 
The metrics used for the evaluation are described in Section \ref{metrics}.
We evaluate the quality of the triplet extraction and salient triplet extraction methods in an automatic evaluation setup.
To do so, we select 50 articles from the Computer Science domain, extract their triplets using different models, and evaluate their quality using GPT4.
We use the same setup to evaluate different salient triplet extraction methods.

%\textbf{Dataset for Evaluating Model}
%\todo{rewrite}

\subsection{Evaluation Criteria}
\label{metrics}
\subsubsection{QA Evaluation}
For evaluating the quality of the generated QAs, we use several metrics inspired by previous work \cite{es2023ragas, yu2024evaluation}. We use the following metrics to evaluate the questions:
\textbf{Relevance} assesses how closely the question aligns with the main concepts of the article (relevant, somewhat relevant, highly relevant); \textbf{Specificity} assesses if the question is broad, vague, or sufficiently detailed for the article content; \textbf{Clarity} assesses how well the question is phrased/structured (low, moderate, or high clarity).

For the answer evaluation, we use the following metrics:
\textbf{Relevance} assesses how well the answer is related to the question and context; \textbf{Factuality} assesses whether the answer is correct based on the contextual information; \textbf{Specificity} assesses whether the answer is overly general or provides specific and detailed information; \textbf{Completeness} assesses whether the answer contains all the necessary information to fully address the question; \textbf{Grammatical Correctness} assesses whether the answer contains grammatically correct and complete sentences; \textbf{Reference Relevance} assesses how well the answer is connected to the reference(s) used to generate the answer.

These evaluation criterion are used both in SME and in automatic evaluation of the QAs.

\subsubsection{Triplet Extraction Metrics}
Once the triplets are extracted using different methods, we use GPT4 to evaluate their quality in two ways: 
\begin{itemize}
    \item \textbf{Cross-model evaluation}: for this evaluation, we use the concept of LLM-as-a-judge \cite{zheng2023judging} and compute the win rate \cite{rafailov2024direct} across models. We ask GPT4 to compare the set of triplets extracted by different models with each other and assign a binary score to each set (0: the other set is better or both sets have a low quality, 1: the set is better than the other or both sets have a high quality). We report \textbf{win rate} as the normalized sum of scores over all articles per model. 
    \item \textbf{Model specific evaluation}: given a list of triplets extracted per model, we ask GPT4 to assign a binary label per triplet reflecting whether the triplet is correctly extracted or not. We then report the average \textbf{accuracy} of the model over all articles.
\end{itemize}
%by comparing the set of triplets extracted by different models with each other. 

\subsubsection{Salient Triplet Extraction Metrics}
We use a methodology similar to the triplet extraction evaluation to evaluate the salient triplet extraction methods. Since we have more than two models in this evaluation, for cross-model evaluation, instead of a binary score, we use a graded scale ($0-2$ as we have three models for this task), sum the scores of models across all articles, and report it as the win rate. For model-specific evaluation, since the task is a ranking task, we report Mean Reciprocal Rank (MRR) per model where the relevance label of triplets at each rank is determined by GPT4.

\subsection{Baselines}
\subsubsection{QA Generation} 
To the best of our knowledge, there are no existing methods specifically designed to generate QAs from full text articles. We use GPT-3.5 as a baseline because it has demonstrated high performance in summarization and question-answering tasks, which are closely related to our objectives. We feed the article (containing all the metadata and the full text) to the model and instruct it to generate QAs. We adapt the same instructions used by the CCQG for this experiment. 

\subsubsection{Triplet Extraction} 
We use the out-of-the-shelf REBEL model as the baseline and compare its performance to that of our fine-tuned REBEL model.

\subsubsection{Salient Triplet Extraction}
We use the following baselines for the salient triplet extraction task:
\begin{itemize}
    \item \textbf{TF}: we only use the frequency of entities in relationships next to the triplet frequency in the article to rank triplets.
    \item \textbf{TF-IDF}: next to the frequency, we use the inverse document frequency of the triplet and its entities to rank triplets.
    %\item \textbf{Semantic Similarity}: in this approach, triplets are ranked by their semantic similarity to the abstract of the article
\end{itemize}

\subsection{Models and Hyperparameters}
We use GPT3.5 for question and answer generation in CCQG and KG-based QA generation pipelines. 
%Next to our two proposed methods we include a baseline method which takes the full-text of an article and directly generates the QAs for an article in a single prompt. 
%The prompt used can also be found in section \ref{sec:promptsappendix}.
%We use the same model to extract triplets from the representative documents in the creation of the dataset to fine-tune the triplet extraction model.
For evaluating the quality of extracted triplets and salient triplets, we use GPT4.
For GPT3.5 and GPT4, we set the temperature of the model to zero. 
For the fine-tuned triplet extraction model, we use a beam size of 2 and a maximum length of 32 for generation.
We finetune a separate REBEL model per science domain and use it to build a KG per domain. To fine-tune the REBEL model, we use a learning rate of $2e-5$, weight decay of $0.01$ and a training batch size of $8$. We train for $10$ epochs and use early stopping.

% \cite{} which is a BART-based model. \todo[inline]{@Zi Long: please add hyperparameters of the model. Zi Long: DONE}
For all similarity estimations (similarity of contexts in the CCQG approach to the paragraphs and similarity of triplets to the triplets extracted from the abstract), we use the \textit{all-mpnet-base-v2} model\footnote{\url{https://huggingface.co/sentence-transformers/all-mpnet-base-v2}}. The threshold for selecting salient paragraphs in the CCQG method is set to 0.7 and the paragraphs with a similarity score higher than this to the proxy are selected as the salient paragraphs.
For similarity estimation in the question ranking component, we use \textit{multi-qa-mpnet-base-cos-v1} model\footnote{\url{https://huggingface.co/sentence-transformers/multi-qa-mpnet-base-cos-v1}}. 
For all QA generation pipelines, we aim to generate at most five QA pairs per article.

\section{Results}
\label{results}
In this section, following our research questions described in Section \ref{experimentalSetup}, we report the results of different QA generation methods. We further evaluate an application of the generated QAs in building scientific question and answer generation models. The details of this experiment can be found in Section \ref{appendix:tuning}.

\begin{table}[!t]
	\centering 
	\caption{\label{table:1}
          Results of CCQG and KG-based QA generation pipelines on question generation.} 
          \vspace{-0.5\baselineskip}
	\begin{tabular}{l l c c c }
        \toprule
        \multicolumn{5}{c}{\textbf{SME Evaluation}}\\
        \hline
        \textbf{Domain}&\textbf{Method} & \textbf{Rel.} & \textbf{Spec.} & \textbf{Clar.} \\
        \hline
        \multirow{2}{2cm}{{Com. Sci.}}& CCQG   & 0.86 & 0.90  & 0.85 \\ 
        & KG     & \textbf{0.93} & \textbf{0.94}  & \textbf{0.91}\\
        \hline
        \multirow{2}{2cm}{{Lif. Sci.}}& CCQG   & 0.83 & 0.81  & 0.82 \\ 
        & KG     & \textbf{0.95} & \textbf{0.94}  & \textbf{0.94}\\
        \hline
        \multicolumn{5}{c}{\textbf{Automatic Evaluation}}\\
        %\hline
        %\textbf{Domain}&\textbf{Method} & \textbf{Rel.} & \textbf{Spec.} & \textbf{Clar.} \\
        \hline
        \multirow{2}{2cm}{{Com. Sci.}}& CCQG   & {0.83} & {0.80}  & {0.90} \\ 
        & KG       & {0.83} & 0.79  & 0.87\\
        \hline
        \multirow{2}{2cm}{{Lif. Sci.}}& CCQG   & {0.83} & {0.84}  & {0.77} \\ 
        & KG       & {0.86} & 0.87  & 0.80\\
        \bottomrule
        
        \end{tabular}
\end{table}

\begin{table*}
	\centering 
	\caption{\label{table:2}
          Results of CCQG and KG-based QA generation pipelines on answer generation.} 
          \vspace{-0.5\baselineskip}
	\begin{tabular}{l l c c c c c c}
        \toprule
        \multicolumn{8}{c}{\textbf{SME Evaluation}}\\
        \hline
        \textbf{Domain} & \textbf{Method} & \textbf{Rel.} & \textbf{Fact.} & \textbf{Spec.} & \textbf{Comp.} & \textbf{Gram. Corr.} & \textbf{Ref. Rel.}\\
        \hline
        \multirow{2}{2cm}{{Com. Sci.}}& CCQG   & 0.87 & 0.89 & 0.85 & 0.86 & 0.96 & 0.87\\
        &KG     & \textbf{0.92} & \textbf{0.92} & \textbf{0.87} & \textbf{0.89} & \textbf{0.99} & \textbf{0.90}\\
        \hline
        \multirow{2}{2cm}{{Lif. Sci.}}& CCQG   & 0.81 & 0.83 & 0.80 & 0.80 & 0.89 & 0.82\\
        &KG     & \textbf{0.95} & \textbf{0.95} & \textbf{0.92} & \textbf{0.92} & \textbf{0.92} & \textbf{0.94}\\
        \hline
        \multicolumn{8}{c}{\textbf{Automatic Evaluation}}\\
        %\hline
        %\textbf{Method} & \textbf{Relevance} & \textbf{Factuality} & \textbf{Specificity} & \textbf{Completeness} & \textbf{Gram. Correct.} & \textbf{Ref. Relevance}\\
        \hline
        \multirow{2}{2cm}{{Com. Sci.}}&CCQG   & 0.79 & 0.81 & 0.82 & 0.77 & 0.99 & 0.80\\
        &KG       & 0.79 & 0.80 & 0.82 & 0.75 & 0.99 & 0.80\\
        \hline
        \multirow{2}{2cm}{{Lif. Sci.}}&CCQG   & 0.78 & 0.80 & 0.79 & 0.76 & 0.99 & 0.78\\
        & KG       & 0.80 & 0.80 & 0.92 & 0.88 & 0.99 & 0.84\\
        \bottomrule
        
        \end{tabular}
\end{table*}

\subsection{Results of different QA generation methods}
\label{resultsRQ1}
To answer our first research question, we evaluate the quality of the generated QAs based on different methods using the evaluation criterion. 
Table \ref{table:1} and Table \ref{table:2} show the quality of the generated questions and answers with respect to the evaluation metrics. 
The results indicate that the KG-based approach achieves a better SME score in terms of all metrics in both domains. 
The question specificity achieves a high score for the KG-based approach, showing the ability of this approach in generating questions with the right amount of detail. 
The answers generated by the KG-based approach receive a high score in terms of relevance and factuality, indicating that the answers address the question well using the contextual paragraphs provided around the salient entities. 

The automatic evaluation does not reflect the differences between different models very well.
It should be noted that the purpose of the automatic evaluation is to establish a lower bound on performance. Therefore, the scores for this evaluation are lower compared to the SME evaluation.
%Therefore, the design of the automatic evaluation aims at establishing that lower bound for the quality of QAs.
%However, based on the automatic evaluation both model achieve good scores which are comparable to the ones achieved by the SME evaluation.
Table \ref{table:3} shows the inter-annotator agreement (using Cohen's Kappa) on the Computer Science's QAs. We measure the agreement between the two SMEs as well as the agreement between the SMEs and the LLM. 
For computing the agreement between the SMEs and the LLM, we compute the agreement between the LLM and each SME on all questions/answers and take the average. 
As the results show, there is high agreement between SMEs and between SMEs and the LLM on both questions and answers. 
Interestingly, the agreement between the SMEs and the LLM is higher than the agreement between SMEs showing that the alignment between the LLM and each individual SME is higher than the alignment between the SMEs. 

\begin{table}
	\centering 
	\caption{\label{table:3}
          Agreement rates between SMEs and between SMEs and the LLM on QA evaluation.} 
          \vspace{-0.5\baselineskip}
	\begin{tabular}{l c c c c}
        \toprule
        \multirow{2}{2cm}{\textbf{Method}} & \multicolumn{2}{c}{SME vs SME} & \multicolumn{2}{c}{SME vs LLM}\\
        \cmidrule(lr){2-3} \cmidrule(lr){4-5}
         & \textbf{Ques.} & \textbf{Ans.} & \textbf{Ques.} &\textbf{Ans.} \\
        \hline 
        CCQG   & 0.68 & 0.74 & 0.74 & 0.75\\ 
        KG     & 0.69 & 0.74 & 0.80 & 0.76\\ 
        \bottomrule
        
        \end{tabular}
\end{table}

\begin{figure}
    \centering
    \includegraphics[width=1\linewidth]{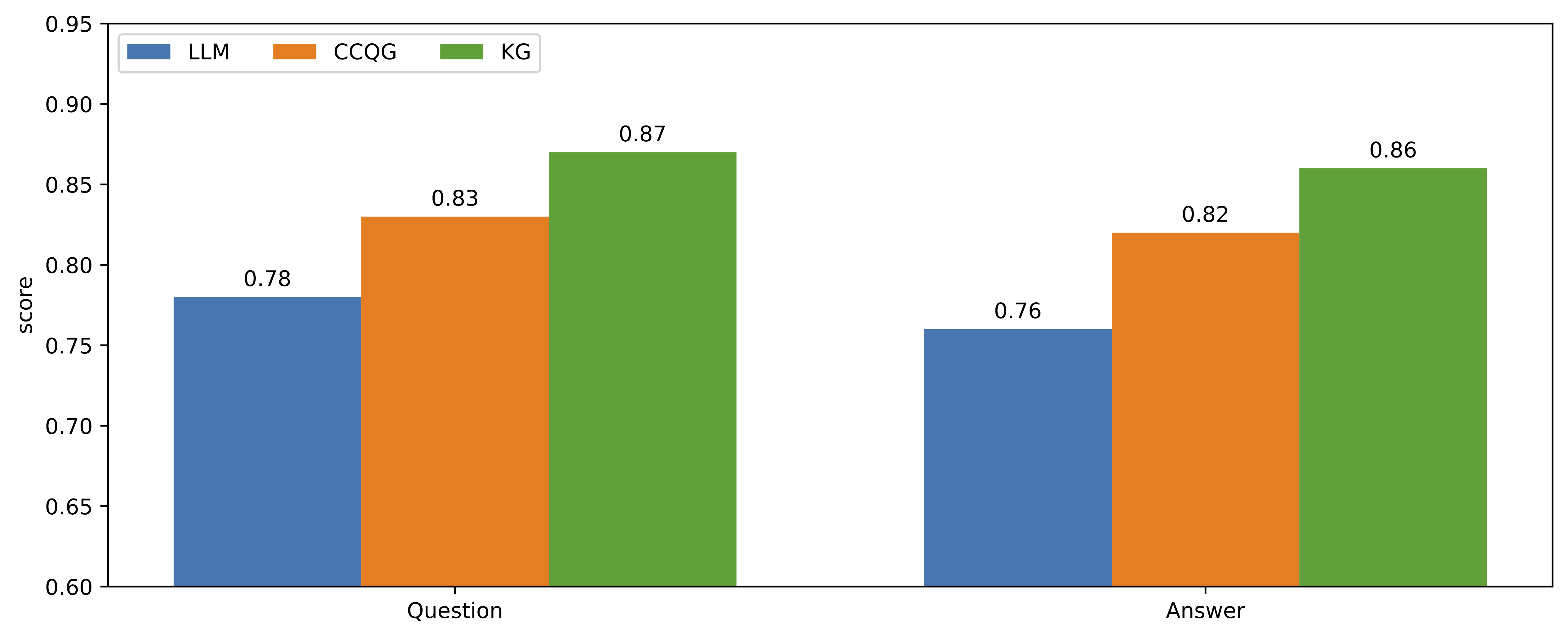}
    \caption{
          Results of CCQG, KG-based, and simple LLM-based QA generation pipelines in terms on win rate.}
    \label{fig:table4}
\end{figure}

As a baseline, we use an LLM (GPT-3.5) to generate QAs given the whole content of the article in one go. 
To do so, we select 50 articles from the Computer Science domain which can fit in a single prompt (together with the instructions). 
We adapt the same instructions used by the CCQG for this experiment. 
After QA generation, we use GPT4 to evaluate the quality of QAs with respect to the evaluation metrics.
Figure \ref{fig:table4} shows the results of this experiment.
For this experiment, we report the average quality scores of questions and answers over all corresponding metrics. 
As the results show, the LLM has the lowest performance compared to the other methods.
This reflects the importance of effective content selection to generate specific and targeted QAs for scientific articles.
Table \ref{table:qa_examples} shows examples of generated QAs using different methods for an article.

\begin{table*}[hbt!]
\centering
\footnotesize
\caption{Top-3 QAs generated by different methods for an article on "Application of data augmentation techniques towards metabolomics". Both KG-based and CCQG methods result in high-quality QAs, however, the questions generated by CCQG method (such as the third question) are more generic and less specific to the article.}
\begin{tabular}{|p{16cm}|}
  \hline
  \textbf{KG Method} \\ 
  \hline
  \RaggedRight \textbf{Q}: "How can data augmentation techniques be applied towards metabolomics?"\newline 
  \textbf{A}: "Data augmentation techniques can be applied towards metabolomics by using computational intelligence-based methods to create more observations in limited and unbalanced metabolomic datasets. These techniques involve the generation of synthetic samples that can be used alongside the original ones for analysis. In the context of metabolomics, data augmentation can help overcome the challenges of imbalanced datasets, especially in rare diseases like Niemann-Pick type C disease. By increasing the number of samples, data augmentation enhances the potential for detecting unusual metabolic patterns and supports the application of multivariate data analysis techniques." \\
  \RaggedRight \textbf{Q}: "What are the limitations of metabolomic datasets?"\newline
  \textbf{A}: "Metabolomic datasets are often limited in the number of samples and heavily unbalanced. Medical datasets, as well as metabolomics ones, may specifically present a characteristic that normally renders the task of prediction/classification difficult: large imbalances amongst the data classes. These sets usually show more samples of the control class than samples belonging to the class that indicates the presence of a particular disease."\\
  \RaggedRight \textbf{Q}: "How can selected metabolites and their concentrations be used to determine the status of different groups of samples?"\newline
  \textbf{A}: "Metabolomics strategies can be used to determine the status of different groups of samples by analyzing bioanalytical dataset systems under pre-established conditions. The concentrations of selected metabolites can be measured in response to specific stimuli, treatments, or exercise regimens. This non-invasive approach allows for the monitoring of patient diagnosis and prognosis, therapeutic management, and even drug development. Abnormal activities in metabolomics pathways can be identified, which may indicate defective enzymes that can be targeted with drugs."\\
  \hline
  \textbf{CCQG Method} \\ 
  \hline
  \RaggedRight \textbf{Q}: "What are the challenges with metabolomic datasets?"\newline
  \textbf{A}: "Metabolomic datasets are often limited in the number of samples and heavily unbalanced. Medical datasets, including metabolomics datasets, often have imbalances among the data classes, with more samples of the control class than samples indicating the presence of a particular disease. This imbalance is especially common in rare or very diseases. To address this challenge, oversampling techniques such as the synthetic minority oversampling technique (SMOTE) can be applied."\\
  \RaggedRight \textbf{Q}: "What is the impact of DA techniques on the prediction performance of the disease?"\newline
  \textbf{A}: "DA techniques can increase the prediction performance of Niemann-Pick Class C1 (NPC1) disease activity in patients. The use of DA techniques can lead to an increase in sensitivity of 20\%-50\% and an increase in the predictive capacity of 0.3 (out of 1) for machine learning models. These techniques allow the identification of urinary metabolomics biomarkers for the diagnosis, pathogenesis, and monitoring of NPC1 disease. The application of DA techniques can also improve disease diagnosis and the search for new metabolomics information and clinically-acceptable biomarkers."\\
  \RaggedRight \textbf{Q}: "What is the purpose of data augmentation (DA) techniques?"\\
  \textbf{A}: "The purpose of data augmentation (DA) techniques is to improve the performance of machine learning models, especially for applications related to problems involving datasets consisting of images. DA techniques are effective in generating synthetic samples that can be used to train the models and enhance their ability to make accurate predictions. However, applying DA to datasets that are not images, signals, or time series is more complex, as it is difficult to evaluate the quality of the generated synthetic data. This is particularly relevant in domains such as genomics or metabolomics."\\
  \hline
  \textbf{Baseline Method} \\ 
  \hline
  \RaggedRight \textbf{Q}: "What is Niemann-Pick type C disease and what causes it?"\newline
  \textbf{A}: "Niemann-Pick type C disease (NPC) is a very rare neurodegenerative lysosomal storage disease caused by mutations in two genes, NPC1 and NPC2."\\
  \RaggedRight \textbf{Q}: "How does metabolomics contribute to the understanding of Niemann-Pick type C disease?"\\
  \textbf{A}: "Metabolomics allows researchers to develop a deep understanding of how the system explored responds to stimuli, aiding in the diagnosis, prognosis, and drug development for conditions like Niemann-Pick type C disease."\\
  \textbf{Q}: "What are some of the clinical features of Niemann-Pick type C disease?"\\
  \textbf{A}: "Niemann-Pick type C disease presents with symptoms such as clumsiness, ataxia, learning difficulties, vertical gaze paralysis, dysphagia, cataplexy, epilepsy, hepatosplenomegaly, and respiratory dysfunction."\\

  \hline
\end{tabular}
\label{table:qa_examples}
\end{table*}

%Examples on generated QAs are shown in Appendix \ref{qa_examples}.

% \begin{table}[t]
% 	\centering 
% 	\caption{\label{table:4}
%           Results of CCQG, KG-based, and simple LLM-based QA generation pipelines.} 
%           \vspace{-0.5\baselineskip}
% 	\begin{tabular}{l c c}
%         \toprule
%         \textbf{Method} & \textbf{Question} & \textbf{Answer} \\
%         \hline
%         LLM    &  0.78 &  0.76\\
%         CCQG   &  0.83 &  0.82\\ 
%         KG     &  \textbf{0.87} &  \textbf{0.86}\\
%         \bottomrule
        
%         \end{tabular}
% \end{table}

\subsection{Quality of the Extracted Triplets}
\label{resultsRQ2}
To address RQ2, we assess the performance of the REBEL model fine-tuned on our Computer Science corpus and compare it with the performance of the off-the-shelf REBEL model. Table \ref{table:5} presents the results of this comparison.
The cross-model evaluation (win rate) reveals that the triplets extracted by the fine-tuned model generally exhibit higher quality than those extracted by the off-the-shelf model for the majority of articles. Furthermore, the model-specific evaluation (accuracy) supports this finding, showing that most individual triplets detected by the fine-tuned model are of high quality.
Notably, the off-the-shelf model correctly detects only 58\% of the triplets in the documents. 
This highlights the importance of fine-tuning the triplet extraction model on the target corpus, particularly when applying these models to a corpus of scientific documents.
This result indicates the effectiveness of our approach in building the training set (by focusing on key representative documents) for fine-tuning the triplet extraction model. 

\begin{table}[t]
	\centering 
	\caption{\label{table:5}
          Results of the REBEL and the fine-tuned REBEL models on the triplet extraction task.} 
          \vspace{-0.5\baselineskip}
	\begin{tabular}{l c c}
        \toprule
        \textbf{Method} & \textbf{Win rate} & \textbf{Accuracy} \\
        \hline
        REBEL              & 26.5  &  0.58\\
        Fine-tuned REBEL   & \textbf{73.5}  &  \textbf{0.78}\\ 
        \bottomrule
        
        \end{tabular}
\end{table}

\subsection{Salient Triplet Extraction Results}
\label{resultsRQ3}
\begin{figure}[t]
    \centering
    \includegraphics[width=1\linewidth]{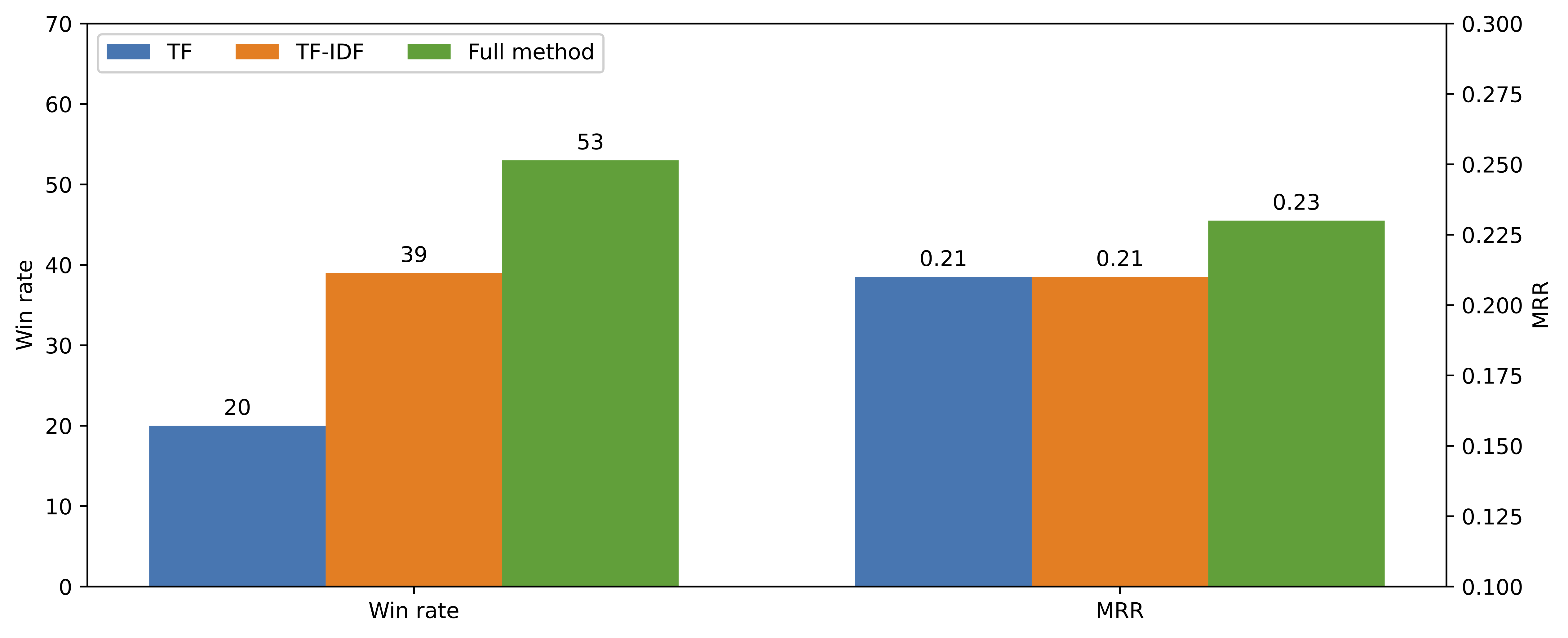}
    \caption{Results of the different salient entity extraction methods in terms of win rate.}
    \label{fig:table6}
\end{figure}

To answer RQ3, we evaluate the performance of the salient triplet extraction method and compare its performance to the baselines in this task. We focus on the model trained on the Computer Science cropus. The results of this experiment are reported in Figure \ref{fig:table6}, which provides a breakdown of the metrics used to assess each method's effectiveness. The full method, which incorporates both frequency and IDF-based saliency measures, demonstrates the highest performance across both model-agnostic and model-specific metrics. 
As expected, the method utilizing triplet IDF outperforms the method that rely solely on triplet frequency. This performance boost underscores the importance of evaluating the saliency of triplets within the broader literature, rather than just their occurrence within a single document. By integrating the triplet's importance as reflected in the entire corpus, the method more accurately identifies the most novel and significant triplets within an article. This finding highlights the necessity of a more nuanced approach to triplet extraction that goes beyond simple frequency counts, ensuring a more accurate and meaningful identification of key triplets.

% Scores dont add up exactly up to 100 because ties are possible
% \begin{table}[t]
% 	\centering 
% 	\caption{\label{table:6}
%           Results of the different salient entity extraction methods.} 
%           \vspace{-0.5\baselineskip}
% 	\begin{tabular}{l c c}
%         \toprule
%         \textbf{Method} & \textbf{Win rate} & \textbf{MRR} \\
%         \hline
%         TF          &  20 &  0.21\\
%         TF-IDF      &  39 &  0.21\\ 
%         Full method &  \textbf{53} &  \textbf{0.23}\\
%         \bottomrule
        
%         \end{tabular}
% \end{table}

\subsection{Fine-tuning LLMs on the QA dataset}
\label{appendix:tuning}
To showcase the applicability of the generated QA dataset in training scientific QA generation models, we fine-tune an LLM on the question generation and another LLM on the answer generation tasks.
For this purpose, we use the prompts, context, and the generated questions and answers from the CCQG method to build two datasets to fine-tune two separate LLMs: the model $\mathcal{M}_Q$ for generating questions and the model $\mathcal{M}_A$ for answering questions.
We take 370k input prompts from the QAs generated for the Computer Science data and their resulting questions. 
We use 80\% as our training set to fine-tune the model $\mathcal{M_Q}$ and use 20\% for evaluation. 
For the answer generation model, $\mathcal{M}_A$, we use 91k QA pairs, which passed the hallucinations checks, where 90k is used for fine-tuning and 1k for evaluation. 

For fine tuning, we use the \textit{zephyr-7b-beta} model\footnote{\url{https://huggingface.co/HuggingFaceH4/zephyr-7b-beta}} as our base LLM for both the $\mathcal{M}_Q$ and $\mathcal{M}_A$ models.

To evaluate the fine-tuned models, we calculate different ROUGE scores between the generated questions and answers and their true counterparts. The results are reported in Table \ref{table:7}. As the results show, both the generated questions and the answers achieve very high ROUGE scores, indicating that they contain the same words mostly in the same order compared to the true values. This reflects that even relatively small LLMs can be fine-tuned on our generated QA datasets for generating scientific QAs.
\begin{table}[t]
	\centering 
	\caption{\label{table:7}
          Results of fine-tuning \textit{zephyr-7b-beta} for question and answer generation. $R$ is short for ROUGE.} 
          \vspace{-0.5\baselineskip}
	\begin{tabular}{l c c c c}
        \toprule
        \textbf{Model} & \textbf{R-1} & \textbf{R-2} & \textbf{R-L} & \textbf{R-LSUM}\\
        \hline
        $\mathcal{M}_Q$     &  0.89 &  0.75 & 0.86 & 0.86\\
        $\mathcal{M}_A$     &  0.96 &  0.93 & 0.96 & 0.96\\ 
        \bottomrule
        
        \end{tabular}
\end{table}

\section{Conclusion}
\label{conclusion}
The rapid growth of scientific content makes it increasingly challenging for researchers to efficiently identify and comprehend the core contributions and ideas within the vast corpus of literature. 
In this paper, we tackled the task of generating Question-Answer (QA) pairs from scientific articles to encapsulate the main ideas of articles in a structured and targeted format.
We proposed two methods for the QA generation task. Our first method uses a proxy to select salient paragraphs, around which the QAs are generated using an LLM. This involves generating questions, post-processing them, ranking them, and finally generating answers. 
Although straightforward and intuitive, this approach lacks a comparison with the existing literature.
To address this, our second approach constructs a Knowledge Graph (KG) from all articles to assess the novelties of individual articles within the context of the broader literature. This method involves fine-tuning a triplet extraction model, building a KG, and developing a method to calculate the saliency of triplets.
We conducted extensive evaluations on the QA generation task, as well as on triplet extraction and salient triplet extraction tasks. 
Our evaluations demonstrate the effectiveness of the KG-based method in generating specific and targeted questions. Comparisons with a strong baseline (GPT-3.5) show that our KG-based method can generate more informative QAs. 

There are multiple possible directions to extend the work done in this paper.
%Our current experiments have been limited to 20,000 articles within the Computer Science domain. 
A possible future work would be to expand our approach to other scientific domains and increase the size of the corpus. 
%This would enable a more comprehensive evaluation of our methods and potentially uncover domain-specific challenges and insights. 
The SME evaluation of the generated QAs has been conducted on a limited sample of 200 QAs. Future work would involve extending this evaluation to a larger and more diverse set of QAs. 
%The QA data can be a valuable resource for further refining question and answer generation models. Future work can focus on using this data to fine-tune existing models
%, improving their ability to generate accurate, relevant, and contextually appropriate QAs. 
%The generated QA pairs have the potential to be utilized for training and evaluating scientific question answering systems. 
%Future research can explore the integration of the generated QAs into existing QA systems, assessing their impact on system performance. 
Future research can explore applications of QA in the development of specialized QA systems capable of handling the complexities of the scientific literature.

%\clearpage
%\input{latex/limitations}

%\appendix
%\input{latex/prompts_appendix}

%\clearpage

\if 0
\bibliographystyle{ACM-Reference-Format}
\bibliography{custom}
\fi

\end{document}